\documentclass[conference]{IEEEtran}
\usepackage{times}

\usepackage[numbers]{natbib}
\usepackage{multicol}
\usepackage[bookmarks=true]{hyperref}
\usepackage{amsmath}
\usepackage{amsthm}
\usepackage{amssymb}
\usepackage{graphicx}
\usepackage[table]{xcolor}
\usepackage{multirow}
\usepackage{color}  
\usepackage[normalem]{ulem}
\usepackage{soul}
\usepackage[ruled]{algorithm2e}

\usepackage{boldline}
\newcounter{RamCount}
\newcounter{DavidCount}
\newcounter{BrentCount}

\newcommand{\Real}{\mathbb{R}}

\begin{document}
\setlength{\textfloatsep}{8pt}

\title{Modeling and Control of Soft Robots Using the Koopman Operator and Model Predictive Control}


\author{
\authorblockN{Daniel Bruder}
\authorblockA{Mechanical Engineering\\
University of Michigan\\
Ann Arbor, Michigan 48109\\
Email: bruderd@umich.edu}
\and
\authorblockN{Brent Gillespie}
\authorblockA{Mechanical Engineering\\
University of Michigan\\
Ann Arbor, Michigan 48109\\
Email: brentg@umich.edu}
\and
\authorblockN{C. David Remy}
\authorblockA{ Mechanical Engineering\\
University of Michigan\\
Ann Arbor, Michigan 48109\\
Email: cdremy@umich.edu}
\and
\authorblockN{Ram Vasudevan}
\authorblockA{Mechanical Engineering\\
University of Michigan\\
Ann Arbor, Michigan 48109\\
Email: ramv@umich.edu} %
}


%

\maketitle

\begin{abstract}
Controlling soft robots with precision is a challenge due in large part to the difficulty of constructing models that are amenable to model-based control design techniques.
Koopman operator theory offers a way to construct explicit linear dynamical models of soft robots and to control them using established model-based linear control methods.
This method is data-driven, yet unlike other data-driven models such as neural networks, it yields an explicit control-oriented linear model rather than just a ``black-box'' input-output mapping.
This work describes this Koopman-based system identification method and its application to model predictive controller design.
A model and MPC controller of a pneumatic soft robot arm is constructed via the method, and its performance is evaluated over several trajectory following tasks in the real-world. 
On all of the tasks, the Koopman-based MPC controller outperforms a benchmark MPC controller based on a linear state-space model of the same system.
\end{abstract}

\IEEEpeerreviewmaketitle

\section{Introduction} 
\label{sec:intro}

Soft robots have bodies made out of intrinsically soft and/or compliant materials.
This inherent softness enables them to safely interact with delicate objects, and to passively adapt their shape to unstructured environments \cite{rus2015design}.
Such traits are desirable for robotic applications that demand safe human-robot interaction such as wearable robots, in-home assistive robots, and medical robots.
Unfortunately, the soft bodies of these robots also impose modeling and control challenges, which have restricted their functionality to date. 
While many novel soft devices such as soft grippers \cite{ilievski2011soft}, crawlers \cite{tolley2014resilient}, and swimmers \cite{marchese2014autonomous} exploit the flexibility of their bodies to achieve coarse behaviors such as grasping and locomotion, they do not exhibit precise control capabilities.



\begin{figure}
    \centering
    \includegraphics[width=\linewidth]{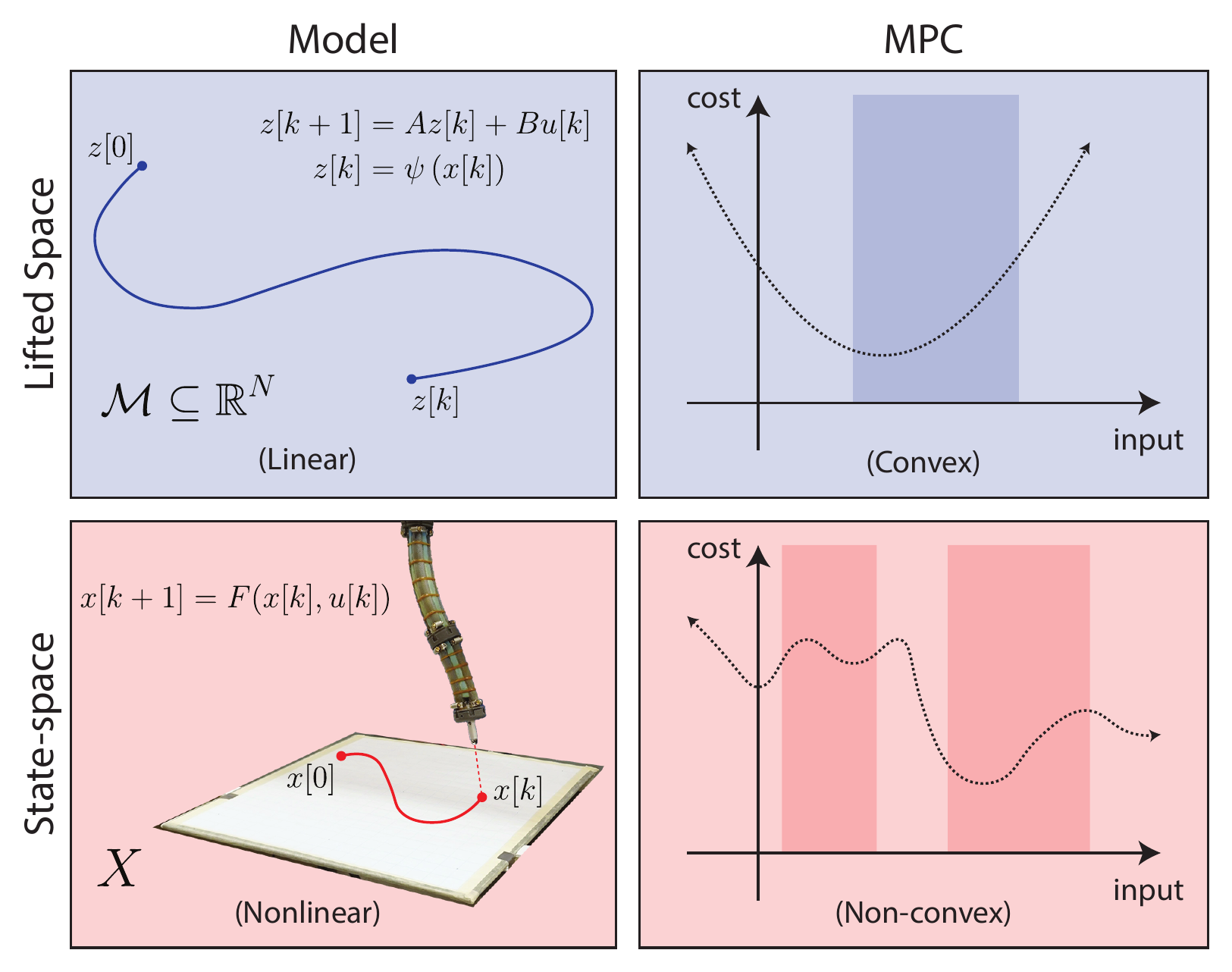}
    \caption{A nonlinear dynamical system (bottom-left) has a \emph{linear} representation in the \emph{lifted} space made up of all real-valued functions (top-left). While a model predictive controller (MPC) designed for the nonlinear system in state-space requires solving a non-convex optimization problem to choose inputs at each time-step (bottom-right), this problem is convex for an MPC controller designed for the lifted linear system (top-right). This paper develops a data-driven method to construct such a lifted model representation for soft robotic systems in the presence of outliers and a to construct a convex, model-based control design technique for such systems. }
    \label{fig:overview}
\end{figure}

The challenge in constructing such precise control techniques is due in large part to the difficulty of devising models of soft robots that are amenable to model-based control design techniques.
Consider for instance a rigid-bodied robotic system that is made up of rigid links connected together by discrete joints.
Since joint displacements can be used to fully describe the configuration of a rigid-bodied system, joint displacements and their derivatives make a natural choice for the state variables for rigid-bodied robots \cite{spong2008robot}.
One can use this choice of state variables to describe the dynamics of the rigid-bodied robot.
This, as a result, makes the application of model-based control design techniques such as feedback linearization \cite{spong2008robot}, nonlinear model predictive control \cite{allgower2012nonlinear}, LQR-trees \cite{tedrake2010lqr}, sequential action control \cite{ansari2016sequential}, and others feasible.

Soft robots, in contrast, do not exhibit localized deformation at discrete joints, but instead deform continuously along their bodies and have infinite degrees-of-freedom.
In the absence of joints, there does not yet exist a canonical choice of state variables to describe the geometry of a soft robot.
As a result, existing representations are typically only rich enough to describe the system under restrictive simplifying assumptions.
For example, the popular piecewise constant curvature model \cite{webster2010design} provides a low-dimensional description of the shape of continuum robots, but only under the assumption that bending occurs in sections of constant curvature.
Other simplified models such as pseudo-rigid-body \cite{howell1996evaluation} and quasi-static \cite{bruder2018iros, thuruthel2018model, gravagne2003large, trivedi2008geometrically} have proven useful,
but they are only able to describe behavior in the subset of conditions over which the simplifying assumptions hold.
This can make applying model-based control design techniques impractical. 

Alternatively, data-driven methods such as traditional machine learning and deep learning can be applied to construct models for soft robots without making structural simplifying assumptions.
Such models provide a ``black-box'' mapping from inputs to outputs and have been shown to predict behavior well across various configurations of soft robots \cite{gillespie2018learning, thuruthel2018model}.
However, since no explicit model is constructed, these methods are also not amenable to existing model-based control design techniques.


Koopman operator theory offers an approach that can overcome the challenges of modeling and controlling soft robots.
The approach leverages the linear structure of the Koopman operator to construct linear models of nonlinear controlled dynamical systems from input-output data \cite{bruder2018nonlinear, mauroy2016linear}, and to control them using established linear control methods \cite{Abraham-RSS-17, korda2018linear}.
In theory, this approach involves \emph{lifting} the state-space to an infinite-dimensional space of scalar functions (referred to as observables), where the flow of such observables along trajectories of the nonlinear dynamical system is described by the \emph{linear} Koopman operator.
In practice, however, it is not feasible to compute an infinite-dimensional operator, so a modified version of the Extended Dynamic Mode Decompostion (EDMD) is employed to compute a finite-dimensional projection of the Koopman operator onto a finite-dimensional subspace of all observables (scalar functions).
This approximation of the Koopman operator describes the evolution of the output variables themselves, provided that they lie within the finite subspace of observables upon which the operator is projected.
Hence, this approach makes it possible to control the output of a nonlinear dynamical system using a controller designed for its linear Koopman representation.

The Koopman approach to modeling and control is well suited for soft robots for several reasons.
Soft robots pose less of a physical threat to themselves or their surroundings when subjected to random control inputs than conventional rigid-bodied robots. 
This makes it possible to safely collect input-output data over a wide range of operating conditions, and to do so in an automated fashion. 
Furthermore, since the Koopman procedure is entirely data-driven, it inherently captures input-output behavior and avoids the ambiguity involved in choosing a discrete set of states for a structure with infinite degrees of freedom.

The work presented here can be considered an extension of the work on Koopman-based modeling and control of \citet{mauroy2016linear} and \citet{korda2018linear}.
The novel contributions of this work, as depicted in Fig. \ref{fig:overview} are:
\begin{enumerate}
    \item An extension to the Koopman system identification procedure described in \cite{mauroy2016linear} to make the resulting Koopman operator both more sparse and less sensitive to outliers and noise in the training data,
    \item The application of this identified Koopman model for model predictive control of a physical soft robotic system.
\end{enumerate}

The rest of this paper is organized as follows:
In Section \ref{sec:sysid} we formally introduce the Koopman operator and describe how it is used to construct linear models of nonlinear dynamical systems. 
In Section \ref{sec:mpc} we describe how the Koopman model can be used to construct a linear model predictive controller (MPC).
In Section \ref{sec:experiments} we describe the soft robot and the set of experiments used to evaluate the performance of a Koopman-based MPC controller.
In Section \ref{sec:conclusion} concluding remarks and perspectives are provided.

\section{Linear System Identification}
\label{sec:sysid}

Any finite-dimensional, Lipschitz continuous nonlinear dynamical system has an equivalent infinite-dimensional linear representation in the space of all real-valued functions of the system's state \cite[Definition 3.3.1]{lasota2013chaos}.
This linear representation, which is called the Koopman operator, describes the flow of functions along trajectories of the system.
While it is not possible to numerically represent the infinite-dimensional Koopman operator, it is possible to represent its projection onto a finite-dimensional subspace as a matrix.
This section shows that for a given choice of basis functions, a \emph{lifted} linear dynamical system model can be extracted directly from the matrix approximation of the Koopman operator.
The remainder of this sections outlines the approach for constructing the Koopman operator approximation and the linear system representation from data.
Section \ref{sec:mpc} illustrates how this model can be incorporated into a model predictive control algorithm.

\subsection{Koopman Representation of a Dynamical System}

Consider a dynamical system
\begin{align}
    \dot{x}(t) &= F (x(t))
    \label{eq:nlsys}
\end{align}
where $x(t) \in X \subset \Real^n$ is the state of the system at time $t \geq 0$, $X$ is a compact subset, and ${F}$ is a continuously differentiable function.
Denote by $\phi(t,x_0)$ the solution to \eqref{eq:nlsys} at time $t$ when beginning with the initial condition $x_0$ at time $0$.
For simplicity, we denote this map, which is referred to as the \emph{flow map}, by $\phi_t (x_0)$ instead of $\phi (t, x_0)$.

The system can be lifted to an infinite dimensional function space $\mathcal{F}$ composed of all square-integrable real-valued functions with compact domain $X \subset \Real^n$.
Elements of $\mathcal{F}$ are called \emph{observables}.
In $\mathcal{F}$, the flow of the system is characterized by the set 
of Koopman operators 
$U_t : \mathcal{F} \to \mathcal{F}$, for each $t \geq 0$,
which describes the evolution of the observables ${f \in \mathcal{F}}$ along the trajectories of the system according to the following definition:
\begin{align}
    U_t f = f \circ \phi_t,      
    \label{eq:koopman}
\end{align}
where $\circ$ indicates function composition.
As desired, $U_t$ is a linear operator even if the system \eqref{eq:nlsys} is nonlinear, since for $f_1, f_2 \in \mathcal{F}$ and $\lambda_1, \lambda_2 \in \Real$
\begin{align}
    \begin{split}
    U_t (\lambda_1 f_1 + \lambda_2 f_2) &= \lambda_1 f_1 \circ \phi_t + \lambda_2 f_2 \circ \phi_t \\
    &= \lambda_1 U_t f_1 + \lambda_2 U_t f_2.
    \end{split}
\end{align}
Thus, the Koopman operator provides a linear representation of the flow of a nonlinear system in the infinite-dimensional space of observables (see Fig. \ref{fig:overview}) \cite{budivsic2012applied}.
Contrast this representation with the one generated by the (nonlinear) flow map that for each $t \geq 0$ describes how the initial condition evolves according to the dynamics of the system.
In particular if one wants to understand the evolution of an initial condition $x_0$ at time $t$ according to \eqref{eq:nlsys}, then one could solve the nonlinear differential equation to generate the flow map. 
On the other hand, one could apply $U_t$ (a linear operator) to the indicator function centered at $x_0$ (i.e. the function that is $1$ at $x_0$ and zero everywhere else) to generate an indicator function centered at the point $\phi_t(x_0)$.

\subsection{Identification of Koopman Operator}
\label{sec:koopid}

Since the Koopman operator is an infinite-dimensional object, it cannot be represented by a finite-dimensional matrix. 
Therefore, we settle for the projection of the Koopman operator onto a finite-dimensional subspace.
Using a modified version of the Extended Dynamic Mode Decomposition (EDMD) algorithm \cite{williams2015data} originally presented in \cite{mauroy2016linear,mauroy2017koopman}, we identify a finite-dimensional approximation of the Koopman operator via linear regression applied to observed data.

Define ${\bar{\mathcal{F}} \subset \mathcal{F}}$ to be the subspace of $\mathcal{F}$ spanned by ${N>n}$ linearly independent basis functions 
${ \{ \psi_i : \Real^n \to \Real \}_{i=1}^N}$.
We denote the image of $\psi_i$ as $ \mathcal{R}_i$ which is equal to ${ \{ w \in \Real | \exists x \in \Real^n \text{ such that } \psi_i(x) = w  \} }$.
For convenience, we assume that the first $n$ basis functions are defined as
\begin{align}
    &\psi_i(x) = x_i
    \label{eq:xinpsi}
\end{align}
where $x_i$ denotes the $i^{\text{th}}$ element of $x$.
Any observable $\bar{f} \in \bar{\mathcal{F}}$ can be expressed as a linear combination of elements of these basis functions
\begin{align}
    \bar{f} &= \theta_1 \psi_1 + \cdots + \theta_N \psi_N
    \label{eq:fexpanded}
\end{align}
where each $\theta_i \in \Real$.
To aid in presentation, we introduce the vector of coefficients ${\theta = [ \theta_1 \,  \cdots \, \theta_N ]^\top}$ and the \emph{lifting function} ${\psi : \Real^n \to \Real^N}$ defined as:
\begin{align}
    \psi(x) &:= \begin{bmatrix} x_i & \cdots & x_n & \psi_{n+1} (x) & \cdots & \psi_N (x) \end{bmatrix}^\top.
    \label{eq:lift}
\end{align}
We denote the image of $\psi$ as $\mathcal{M} = \mathcal{R}_1 \times \cdots \times \mathcal{R}_N \subset \Real^N$.
By \eqref{eq:fexpanded} and \eqref{eq:lift}, $\bar{f}$ evaluated at a point $x$ in the state space is given by
\begin{align}
    \bar{f}(x) &= \theta^\top \psi (x)
    \label{eq:fvec}
\end{align}
We therefore refer to $\psi(x)$ as the \emph{lifted state}, and $\theta$ as the \emph{vector representation} of $\bar{f}$.

Given this vector representation for observables, a linear operator $L : \bar{\mathcal{F}} \to \bar{\mathcal{F}}$ can be represented as an ${N \times N}$ matrix. 
We denote by $\bar{U}_t \in \Real^{N \times N}$ the approximation of the Koopman operator in $\bar{\mathcal{F}}$, which operates on observables via matrix multiplication:
\begin{align}
    \bar{U}_t \theta = \theta'
\end{align}
where $\theta , \theta'$ are each vector representations of observables in $\bar{\mathcal{F}}$.
Our goal is to find a $\bar{U}_t$ that describes the action of the infinite dimensional Koopman operator $U_t$ as accurately as possible in the $L^2$-norm sense on the finite dimensional subspace $\bar{\mathcal{F}}$ of all observables.

To perfectly mimic the action of $U_t$ on an observable ${\bar{f} \in \bar{\mathcal{F}} \subset \mathcal{F}}$, according to \eqref{eq:koopman} the following should be true for all  $x \in X$
\begin{align}
    \bar{U}_t \bar{f}(x) &= \bar{f} \circ \phi_t(x) \\
    ( \bar{U}_t {\theta} )^\top {\psi}(x) &=
    {\theta}^\top {\psi} \circ \phi_t(x) \\
    \bar{U}_t^\top \psi(x) &= {\psi} \circ \phi_t(x),
    \label{eq:UbarEq}
\end{align}
where the second equation follows by substituting \eqref{eq:fexpanded} and the last equation follows by cancelling $\theta^\top$.
Since this is a linear equation, it follows that for a given ${x \in X}$, solving \eqref{eq:UbarEq} for $\bar{U}_t$ yields the best approximation of $U_t$ on $\bar{\mathcal{F}}$ in the $L^2$-norm sense \cite{penrose1956best}:
\begin{align}
    \bar{U}_t = \left( {\psi}^\top(x) \right)^\dagger ( {\psi} \circ \phi_t(x) )^\top
    \label{eq:Uapprox}
\end{align}
where superscript $\dagger$ denotes the least-squares pseudoinverse.

To approximate the Koopman operator from a set of experimental data, we take $K$ discrete state measurements in the form of so-called ``snapshot pairs'' $(a[k] , b[k])$ for each $k \in \{1,\ldots,K\}$ where
\begin{align}
    a[k] &= x[k] \\
    b[k] &= \phi_{T_s} (x[k]) + \sigma[k],
    \label{eq:ab}
\end{align}
where $\sigma[k]$ denotes measurement noise, $T_s$ is the sampling period which is assumed to be identical for all snapshot pairs, and $x[k]$ denotes the measured state corresponding to the $k^\text{th}$ measurement.
Note that consecutive snapshot pairs do not have to be generated by consecutive state measurements. 
We then lift all of the snapshot pairs according to \eqref{eq:lift} and compile them into the following ${K \times N}$ matrices:
\begin{align}
    &\Psi_a := \begin{bmatrix} {\psi}(a[1])^\top \\ \vdots \\  {\psi}(a[K])^\top \end{bmatrix}
    &&\Psi_b := \begin{bmatrix} {\psi}(b[1])^\top \\ \vdots \\  {\psi}(b[K])^\top \end{bmatrix}
    \label{eq:Psi}
\end{align}
$\bar{U}_{T_s}$ is chosen so that it yields the least-squares best fit to all of the observed data, which, following from \eqref{eq:Uapprox}, is given by 
\begin{align}
    \bar{U}_{T_s} &:= \Psi_a^\dagger \Psi_b.
\end{align}

Sometimes a more accurate model can be attained by incorporating delays into the set of snapshot pairs. 
To incorporate these delays, we define the snapshot pairs as
\begin{align}
    a[k] &= \begin{bmatrix} x[k]^\top, & x[k-1]^\top & \ldots, & x[k-d]^\top \end{bmatrix}^\top \label{eq:snapd1} \\
    b[k] &= \begin{bmatrix} \left( \phi_{T_s} (x[k]) + \sigma_k \right)^\top & x[k]^\top & \ldots & x[k-d+1]^\top \end{bmatrix}^\top \label{eq:snapd2}
\end{align}
where $d$ is the number of delays.
We then modify the domain of the lifting function such that $\psi : \Real^{n+nd} \to \Real^{N}$ to accommodate the larger dimension of the snapshot pairs.
Once these snapshot pairs have been assembled, the model identification procedure is identical to the case without delays.

\subsection{Building Linear System from Koopman Operator}

For dynamical systems with inputs, we are interested in using the Koopman operator to construct discrete linear models of the following form
\begin{equation}
\begin{aligned}
    z[j+1] &= A z[j] + B u[j] \\
    x[j] &= C z[j]
    \label{eq:linSys}
\end{aligned}
\end{equation}
for each $j \in \mathbb{N}$, where $x[0]$ is the initial condition in state space, $z[0] = \psi(x[0])$ is the initial lifted state, $u[j] \in \Real^m$ is the input at the $j^{\text{th}}$ step, and $C$ acts as a projection operator from the lifted space onto the state-space.
Specifically, we desire a representation in which (non-lifted) inputs appear \emph{linearly}, because models of this form are amenable to real-time, convex optimization techniques for feedback control design, as we describe in Section \ref{sec:mpc}.

We construct a model of this form by first applying the system identification method of Section \ref{sec:koopid} to the following modified snapshot pairs
\begin{align}
    &\alpha[k] = \begin{bmatrix} \psi(a[k]) \\ u[k] \end{bmatrix} 
    &&\beta[k] = \begin{bmatrix} \psi(b[k]) \\ u[k] \end{bmatrix}
    \label{eq:alpha}
\end{align}
for each $k \in \{1,\ldots,K\}$. 
The input $u[k]$ in snapshot $k$ is not lifted to ensure that it appears linearly in the resulting model.
With these pairs, we define the following ${K \times (N + m)}$ matrices:
\begin{align}
    &\Gamma_\alpha = \begin{bmatrix} \alpha[1]^\top \\ \vdots \\  \alpha[K]^\top \end{bmatrix}
    &&\Gamma_\beta = \begin{bmatrix} \beta[1]^\top \\ \vdots \\  \beta[K]^\top \end{bmatrix}
    \label{eq:Gamma}
\end{align}
and solve for the corresponding Koopman operator according to \eqref{eq:Uapprox}
\begin{align}
    \bar{U}_{T_s} &:= \Gamma_{\alpha}^\dagger \Gamma_\beta.
    \label{eq:koopGamma}
\end{align}
Note that by \eqref{eq:UbarEq} and \eqref{eq:koopGamma} the transpose of this Koopman matrix is the best approximation of a transition matrix between the elements of snapshot pairs in the $L^2$-norm sense \
\begin{align}
    \bar{U}_{T_s}^\top 
    \begin{bmatrix} \psi(a[k]) \\ u[k] \end{bmatrix} &\approx
    \begin{bmatrix} \psi(b[k]) \\ u[k] \end{bmatrix},
\end{align}
and we desire the best $A,B$ matrices such that
\begin{align}
    A \psi(a[k]) + B u[k] &\approx \psi(b[k])
    \label{eq:linSys_psi}
\end{align}
Therefore, the best $A$ and $B$ matrices of \eqref{eq:linSys} are embedded in $\bar{U}_{T_s}^\top$ and can be isolated by partitioning it as follows:
\begin{align}
    \bar{U}_{T_s}^\top &= 
    \begin{bmatrix} 
        A_{N \times N} &
        B_{N \times m} \\
        O_{m \times N} &
        I_{m \times m}
    \end{bmatrix}
    \label{eq:AB}
\end{align}
where $I$ denotes an identity matrix, $O$ denotes a zero matrix, and the subscripts denote the dimensions of each matrix.
The $C$ matrix is defined
\begin{align}
    C &= \begin{bmatrix} I_{n \times n} & O_{n \times (N-n)} \end{bmatrix}
    \label{eq:C}
\end{align}
since by \eqref{eq:xinpsi}, ${x = [ \psi_1(x) , \dots , \psi_n(x) ]}$.
Note we can also incorporate input delays into the model by appending them to the snapshot pairs as we did in \eqref{eq:snapd1} and \eqref{eq:snapd2}.



\subsection{Practical Considerations: Overfitting and Sparsity} \label{subsec:sparsity}

\begin{figure}
    \centering
    \includegraphics[width=0.8\linewidth]{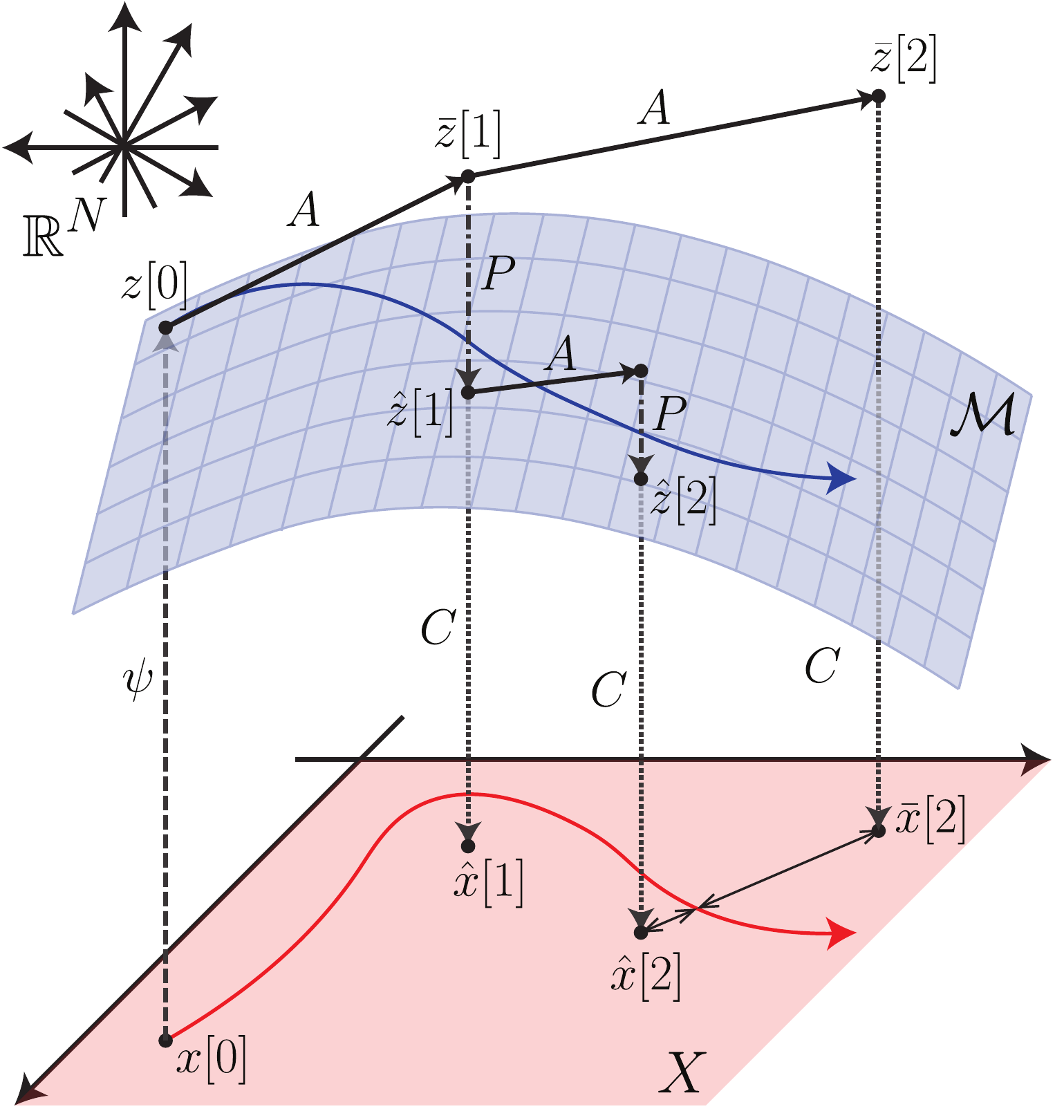}
    \caption{An illustration of the effect of deviating from the image of the lifted functions ${\cal M}$ and how it can be remedied by defining a projection operation as described in Section \ref{subsec:sparsity}. 
    The evolution of the finite dimensional system in the state space $X$ from $x_0$ is depicted as a red curve. 
    The lifted version of this evolution is depicted as the blue curve which is contained in $\cal{M}$.
    The discrete time system representation in the higher-dimensional space created by iteratively applying the state matrix $A$ to $z[j]$ may generate a solution that is outside of ${\cal M}$.
    Though one can still apply $C$ to $\bar{z}$ to project it back to $X$, this may result in poor performance.
    Instead, by projecting $\bar{z}[j]$ onto the manifold at each discrete time step to define a new lifted state $\hat{z}[j]$, the deviation from $\cal{M}$ is reduced, which improves overall predictive performance. }
    \label{fig:manifold}
\end{figure}

A pitfall of data-driven modeling approaches is the tendency to overfit.
While least-squares regression yields a solution that minimizes the total $L^2$ error with respect to the training data, this solution can be particularly susceptible to outliers and noise \cite{rousseeuw2005robust}.
To guard against overfitting to noise while identifying $\bar{U}_{T_s}$, we utilize the $L^1$-regularization method of Least Absolute Shrinkage and Selection Operator (LASSO) \cite{tibshirani1996regression}:
\begin{equation}
\begin{aligned}
\hat{\vec{U}}_{T_s} &= 
& \text{arg}~\underset{ \vec{U}_{T_s} }{\text{min}}
& & || \vec{\Gamma}_\alpha \vec{U}_{T_s} - \vec{\Gamma}_\beta ||_2^2 + \lambda || \vec{U}_{T_s} ||_1
\label{eq:lasso}
\end{aligned}
\end{equation}
where $\lambda \in \Real^{+}$ is the weight of the $L^1$ penalty term, and $\vec{\cdot}$ denotes a vectorized version of each matrix with dimensions consistent with the stated problem.
For $\lambda = 0$, \eqref{eq:lasso} provides the same unique least-squares solution as \eqref{eq:koopGamma}; as $\lambda$ increases it drives the elements of $\vec{U}_{T_s}$ to zero.
For an overview of the LASSO method and its implementation see \citet{tibshirani1996regression}.

The benefit of using $L^1$-regularization to reduce overfitting rather than $L^2$-regularization (e.g. ridge regression) is its ability to drive elements to zero, rather than just making them small.
This promotes sparsity in the resulting Koopman operator matrix (and consequently the $A$ and $B$ matrices).
Sparsity is desirable since it reduces the memory needed to store these matrices on a computer, enabling a higher dimensional set of basis functions to be used to construct the lifting function $\psi$.

Though sparsity is desirable, it can come at the loss of accuracy in prediction. 
As illustrated in Fig. \ref{fig:manifold}, the lifting function $\psi$ maps from $\Real^n$ to $\mathcal{M}$, but at some time step $j$, $A\psi(a[j]) + B u[j]$ may not map onto $\mathcal{M}$.
When this happens and we try to simulate our linear model from an initial condition, it may leave the space of legitimate ``lifted states'' rapidly and fail to predict behavior accurately.
We therefore desire the sparsest model that minimizes the distance from $\mathcal{M}$ at each iteration.

This can be accomplished by applying a projection operator at each time step.
For each snapshot pair, the ideal projection operator $P$ should satisfy the following for all $k$
\begin{align}
    P \left( A {\psi}(a[k]) + B u[k] \right) &= \psi(b[k]).
\end{align}
To build an approximation to this operator, we construct the following $K \times N$ matrix,
\begin{align}
    &\Omega_a := \begin{bmatrix} \left( A {\psi}(a[1]) + B u[1] \right)^\top \\ \vdots \\  \left( A {\psi}(a[K]) + B u[K] \right)^\top \end{bmatrix}.
    \label{eq:Omega}
\end{align}
Then the best projection operator in the $L^2$-norm sense based on our data is given by
\begin{align}
    P := \left( \Omega_{a}^\dagger \Psi_b \right)^\top.
    \label{eq:P}
\end{align}
Composing $P$ with the $A$ and $B$ matrices in \eqref{eq:linSys} yields a modified linear model that significantly reduces the distance from $\mathcal{M}$ at each iteration,
\begin{align}
    z[j+1] &= \hat{A} z[j] + \hat{B} u[j]
    \label{eq:linSys_wP}
\end{align}
where $\hat{A} := PA$ and  $\hat{B} := PB$.
Algorithm \ref{alg:id} summarizes the proposed model construction process.

\begin{algorithm}[t]
\SetAlgoLined
\KwIn{ $\lambda$ , $\{ a[k] , b[k] \}$ and ${ u[k] }$ for $k = 1 , ... , K$}
\textbf{Step 1:} Lift data via \eqref{eq:lift} \\
\textbf{Step 2:} Combine lifted data and inputs via \eqref{eq:alpha} \\
\textbf{Step 3:} Approximate Koopman operator $\bar{U}_{T_s}$ via \eqref{eq:lasso} \\
\textbf{Step 4:} Extract model matrices $A,B$ via \eqref{eq:AB} \\
\textbf{Step 5:} Identify projection operator $P$ via \eqref{eq:P} \\
\KwOut{$\hat{A} := PA$, $\hat{B} := PB$   }
 \caption{Koopman Linear System Identification}
 \label{alg:id}
\end{algorithm}
\section{Model Predictive Control}
\label{sec:mpc}

A system model enables the design of model-based controllers that leverage model predictions to choose suitable control inputs for a given task.
In particular, model-based controllers can anticipate future events, allowing them to optimally choose control inputs over a finite time horizon.
The most popular model-based control design technique is model predictive control (MPC), wherein one optimizes the control input over a finite time horizon, applies that input for a single timestep, and then optimizes again, repeatedly \cite{rawlings2009model}.
For linear systems, MPC consists of iteratively solving a convex quadratic program.

Importantly, this is also the case for Koopman-based MPC control, wherein one solves the following program at each time instance $k$ of the closed-loop operation:
\begin{equation}
\begin{aligned}
& \underset{u[i] , z[i]}{\text{min}}
& & z[N_h]^{T} G_{N_h} z[N_h] + g_{N_h}^T z[N_h] +  \\
&&& + \sum_{i=0}^{N_h - 1} z[i]^T G_i z[i] + u[i]^T H_i u[i] + g_i^T z[i] + h_i^T u[i]\\
& \text{s.t.}
& & z[i+1] = \hat{A} z[i] + \hat{B} u[i] , \quad \forall i \in \{ 0 , \ldots , N_h - 1 \} \\
&&& E_i z[i] + F_i u[i] \leq b_i , \quad \forall i \in \{ 0 , \ldots , N_h - 1\} \\
&&& z[0] = \psi (x[k])
\end{aligned} \label{eq:mpc}
\end{equation}
where $N_h \in \mathbb{N}$ is the prediction horizon, $G_i \in \Real^{N \times N}$ and $H_i \in \Real^{m \times m}$ are positive semidefinite matrices, and where each time the program is called, the predictions are initialized from the current lifted state $\psi (x[k])$.
The matrices $E_i \in \Real^{c \times N}$ and $F_i \in \Real^{c \times m}$ and the vector $b_i \in \Real^{c}$ define state and input polyhedral constraints where $c$ denotes the number of imposed constraints.
While the size of the cost and constraint matrices in \eqref{eq:mpc} depend on the dimension of the lifted state $N$, \citet{korda2018linear} show these can be rendered independent of $N$ by transforming the problem into its so-called ''dense-form.''
Algorithm \ref{alg:mpc} summarizes the closed-loop operation of this Koopman based MPC controller.

\begin{algorithm}[t]
\SetAlgoLined
\KwIn{ Prediction horizon: $N_h$ \\
       \hspace{32pt} Cost matrices: $G_i , H_i , g_i , h_i$ for  $i = 0 , ... ,N_h$ \\
       \hspace{32pt} Constraint matrices: $E_i , F_i , b_i$ for $i = 0 , ... ,N_h$ \\
       \hspace{32pt} Model matrices: $\hat{A} , \hat{B}$ }
\For{ $k = 0 , 1 , 2 , ... $}{
\textbf{Step 1:} Set $z[0] = \psi ( x[k] )$ \\
\textbf{Step 2:} Solve \eqref{eq:mpc} to find optimal input $(u[i]^*)_{i=0}^{N_h}$ \\
\textbf{Step 3:} Set $u[k] = u[0]^*$ \\
\textbf{Step 4:} Apply $u[k]$ to the system
}
 \caption{Koopman-Based MPC}
 \label{alg:mpc}
\end{algorithm}


Since this optimization problem is convex, it has a unique globally optimal solution that can efficiently be constructed without initialization \cite{boyd2004convex} for models with thousands of states and inputs  \cite{paulson2014fast,stellato2018osqp}.
This contrasts sharply with the MPC formulation for nonlinear systems (referred to as nonlinear model predictive control or NMPC \cite{allgower2012nonlinear}).
NMPC requires solving an optimization problem with nonlinear constraints and a (potentially) nonlinear cost function.
As a result, algorithms to solve such problems typically require initialization and can struggle to find globally optimal solutions \cite{polak2012optimization}.
Though techniques have been proposed to improve the speed of algorithms to solve NMPC problems \cite{patterson2014gpops,hereid2017frost} or even globally solve such problems without requiring initialization \cite{zhao2017control}, these formulations still take several seconds per iteration, which can make them too slow to be applied during real-time control.

\section{Experiments}
\label{sec:experiments}

This section describes the robot and the set of experiments used to demonstrate the efficacy of the modeling and control methods from Sections \ref{sec:sysid} and \ref{sec:mpc}.
Video of the soft robot performing several tasks from the final experiment is included in a supplementary video file\footnote{https://youtu.be/e35o2OPsQHs}.

\subsection{Robot Description: Soft Arm with Laser Pointer}
\label{sec:robot}

\begin{figure}
    \centering
    \includegraphics[width=\linewidth]{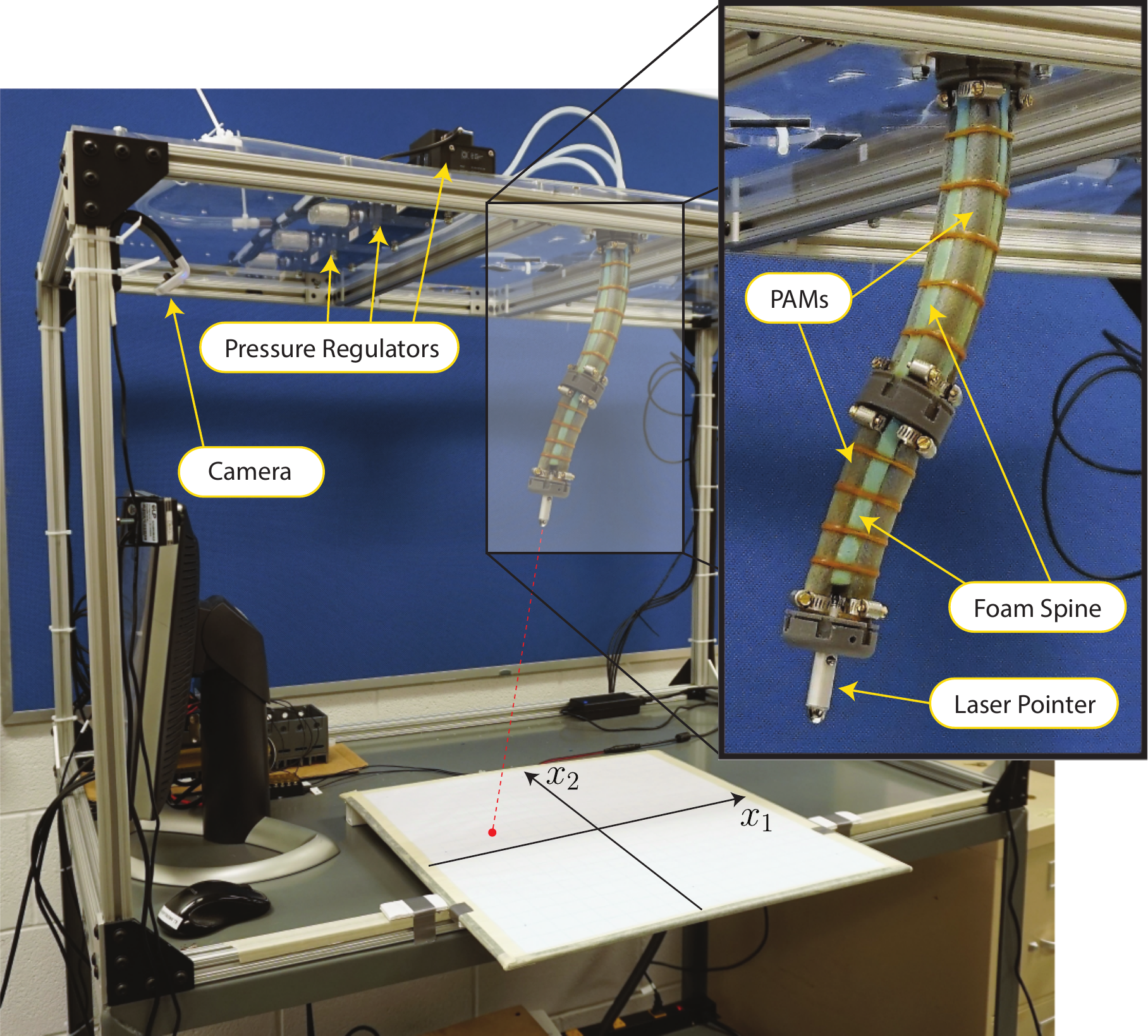}
    \caption{The soft robot consists of two bending segments with a laser pointer attached to the end effector. A set of three pressure regulators is used to control the pressure inside of the pneumatic actuators (PAMs), and a camera is used to track the position of the laser dot.}
    \label{fig:rig}
\end{figure}

The robot used for the experiments is a suspended soft arm with a laser pointer attached to the end effector (see Fig. \ref{fig:rig}). 
The laser dot is projected onto a ${50\text{ cm} \times 50\text{ cm}}$ flat board which sits $34\text{ cm}$ beneath the tip of the laser pointer when the robot is in its relaxed position (i.e. hanging straight down).
The position of the laser dot is measured by a digital webcam overlooking the board.

The arm itself consists of two sections that are each composed of three pneumatic artificial muscles or PAMs (also known as McKibben actuators \cite{tondu2012modelling}) adhered to a central foam spine by latex rubber bands (see Fig. \ref{fig:rig}).
The PAMs in the upper and lower sections are internally connected so that only three input pressure lines are required, and
they are arranged such that for any bending of the upper section, bending in the opposite direction occurs in the bottom section.
This ensures that the laser pointer mounted to the end effector points approximately vertically downward so that the laser light strikes the board at all times.
The pressures inside the actuators are regulated by three Enfield TR-010-g10-s pneumatic pressure regulators, which accept ${0-10}$V command signals corresponding to pressures of ${ \approx 0 - 140 }$ kPa.
In the experiments there is a three-dimensional input corresponding to the voltages into the three pressure regulators and a two dimensional state corresponding to the position of the laser dot with respect to the center of the board.

\subsection{Characterization of Stochastic Behavior}
\label{sec:noise}

Most mechanical systems demonstrate stochastic behavior (i.e. when an identical input and state produces a different output) to some extent.
Stochastic behavior is characteristic of electronic pressure regulators, which can limit the precision of pneumatically driven soft robotic systems
and undermine the predictive capability of models.

\begin{figure}
    \centering
    \includegraphics[width=\linewidth]{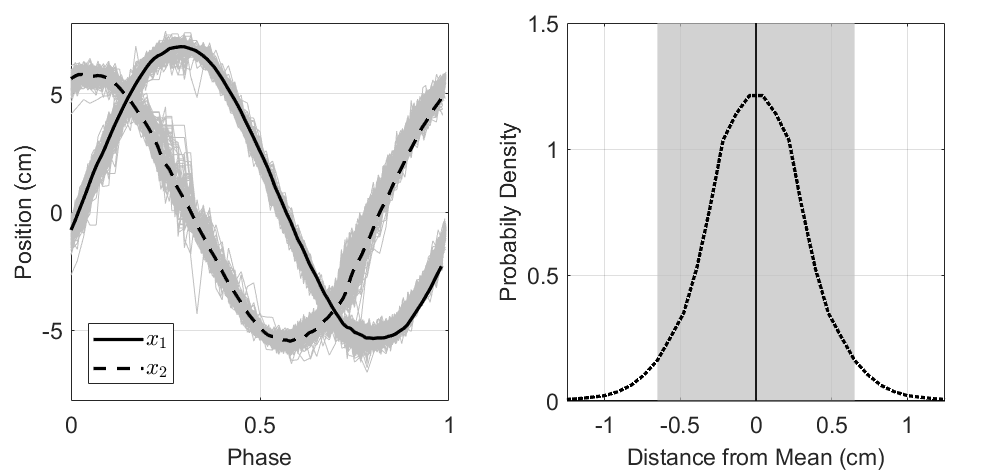}
    \caption{The left plot shows the average response of the system over a single period when the sinusoidal inputs of varying frequencies described by \eqref{eq:unoise} are applied. All of the particular responses are subimposed in light grey.
    The right plot shows the distribution of trajectories about the mean, with all distances within two standard deviations highlighted in grey. The width of the distribution illustrates how for the soft robot system identical inputs can produce outputs that vary by up to 2 cm.}
    \label{fig:noise}
\end{figure}

We quantified the stochastic behavior of our soft robot system by observing the variations in output from period-to-period under sinusoidal inputs to the three actuators of the form
\begin{align} 
    u[k] &= \begin{bmatrix} 6 \sin ( \frac{2 \pi}{T} k T_s ) + 3 \vspace{5pt} \\ 
    6 \sin ( \frac{2 \pi}{T} k T_s - \frac{T}{3} ) + 3 \vspace{5pt} \\ 
    6 \sin ( \frac{2 \pi}{T} k T_s  - \frac{2T}{3}) + 3\end{bmatrix}
    \label{eq:unoise}
\end{align}
for periods of $T = 6,7,8,9,10,11,12$ seconds and a sampling time of $T_s = 0.1$ seconds with a zero-order-hold between samples. 
Under these inputs, the laser dot traces out a circle with some variability in the trajectory over each period.
In Fig.~\ref{fig:noise} the trajectories over 210 periods are superimposed along with the average over all trials.
Nearly all of the observed points fell within 1 cm of the mean trajectory.
Given this inherent stochasticity of our soft robotic system, in the best case we expect only to be able to control the output to within $\approx 1$ cm of a desired trajectory.

\subsection{Data Collection and Model Identification}
\label{sec:datacollection}

To construct a model, we ran the system through $16$ trials each lasting approximately $20$ minutes.
A randomized input was applied during each trial to generate a representative sampling of the system's behavior over its entire operating range.
To ensure randomization, a matrix $\Upsilon \in [0,10]^{3\times 1000}$ of uniformly distributed random numbers between zero and ten was generated to be used as an input lookup table.
Each control input was smoothly varied between elements in consecutive columns of the table over a transition period $T_u$, with a time offset of $T_u / 3$ between each of the three control signals
\begin{align}
    u_i (t) &= \frac{(\Upsilon_{i,k+1} - \Upsilon_{i,k})}{T_u} \left( t + \frac{(i-1) T_u}{3} \right) + \Upsilon_{i,k}
    \label{eq:input}
\end{align}
where $k = \text{floor}\left( {t} / {T_u} \right)$ is the current index into the lookup table at time $t$. 
The transition period $T_u$ varied from $5$ seconds to $10$ seconds between trials.
After collection, the data was uniformly sampled with period $T_s = 0.1$ seconds.

\begin{figure}
    \centering
    \includegraphics[width=\linewidth]{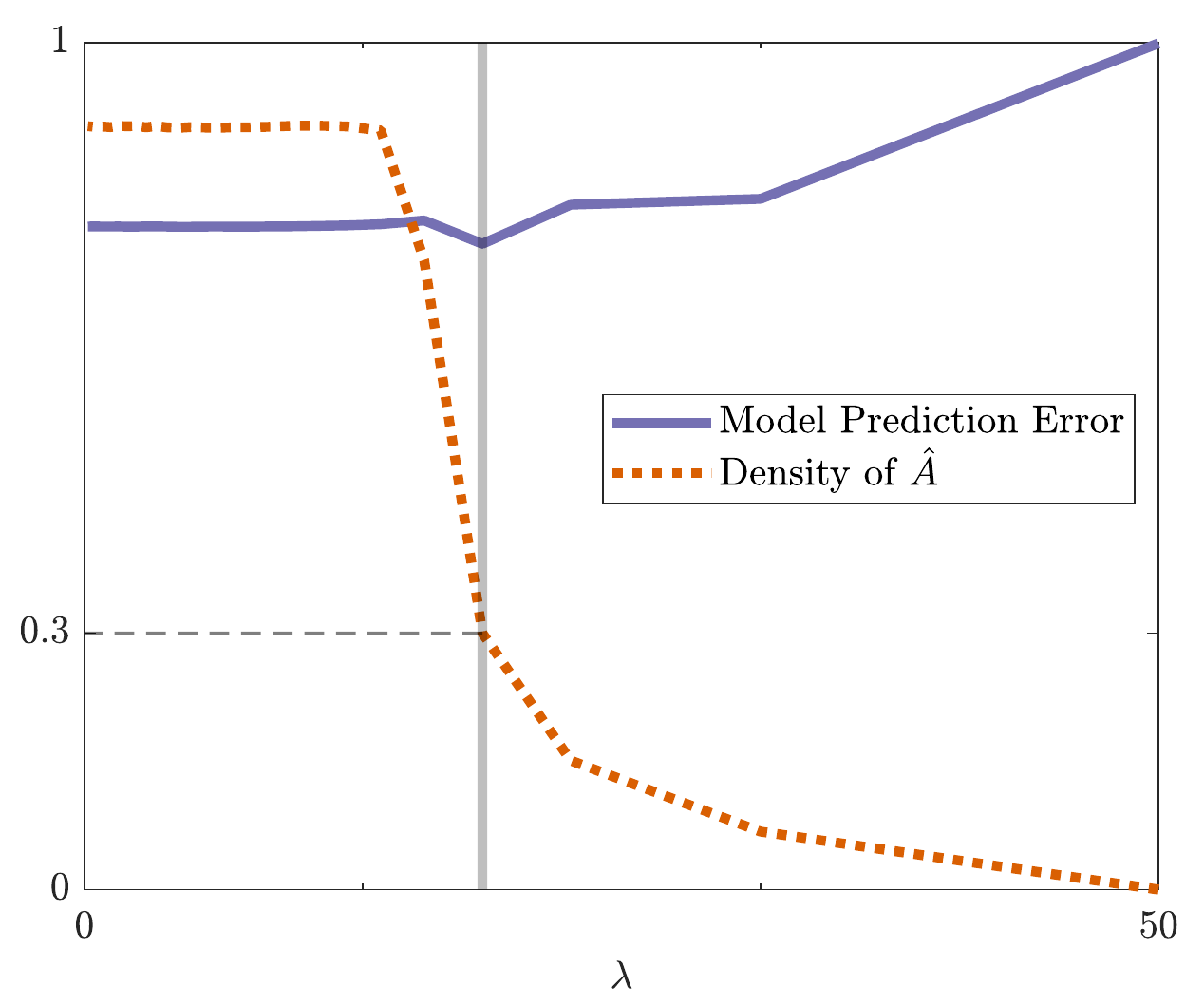}
    \caption{As $\lambda$ (the weight of the $L^1$ penalty term in \eqref{eq:lasso}) increases, the density of the lifted system matrix $\hat{A}$ decreases.
    The model generated by solving \eqref{eq:lasso} with the value of $\lambda$ designated by the vertical grey bar has lower error and a sparser $\hat{A}$ matrix than the least-squares solution to \eqref{eq:koopGamma}, which occurs at $\lambda = 0$.}
    \label{fig:lasso}
\end{figure}

Two models were fit from the data: a Koopman model, and a linear state space model.
The linear state-space model provides a baseline for comparison and was identified from the same data as the Koopman model using the MATLAB System Identification Toolbox \cite{MATLAB:2017}.
This model is a four dimensional linear state-space model expressed in observer canonical form.
The Koopman model was identified via the method described in Section \ref{sec:sysid} on a set of $191,000$ snapshot pairs $\{ a[k] , b[k] \}$ that incorporate a single delay $d = 1$:
\begin{align}
    a[k] &= \begin{bmatrix} x[k]^\top & x[k-1]^\top & u[k-1]^\top \end{bmatrix}^\top \\
    b[k] &= \begin{bmatrix} \left( \phi_{T_s} (x[k]) + \sigma[k] \right)^\top & x[k]^\top & u[k]^\top \end{bmatrix}^\top,
\end{align}
and using an $N=330$ dimensional set of basis functions consisting of all monomials of maximum degree 4.
To find the sparsest acceptable matrix representation of the Koopman operator, equation \eqref{eq:lasso} was solved for ${ \lambda = 0,1,2, ... ,50 }$.
Predictions from the resulting models were evaluated against a subset of the training data, with the error quantified as the average Euclidean distance between the prediction and actual trajectory at each point, normalized by dividing by the average Euclidean distance between the actual trajectory and the origin.
Fig.~\ref{fig:lasso} shows that as $\lambda$ increases so does this error, but the density of the $\hat{A}$ matrix of the lifted linear model decreases.
The model chosen is the one that minimizes prediction error, which results in an $\hat{A}$ matrix with $70 \%$ of its entries equal to zero.

\subsection{Experiment 1: Model Prediction Comparison}
\label{sec:predict}

\begin{figure}
    \centering
    \includegraphics[width=\linewidth]{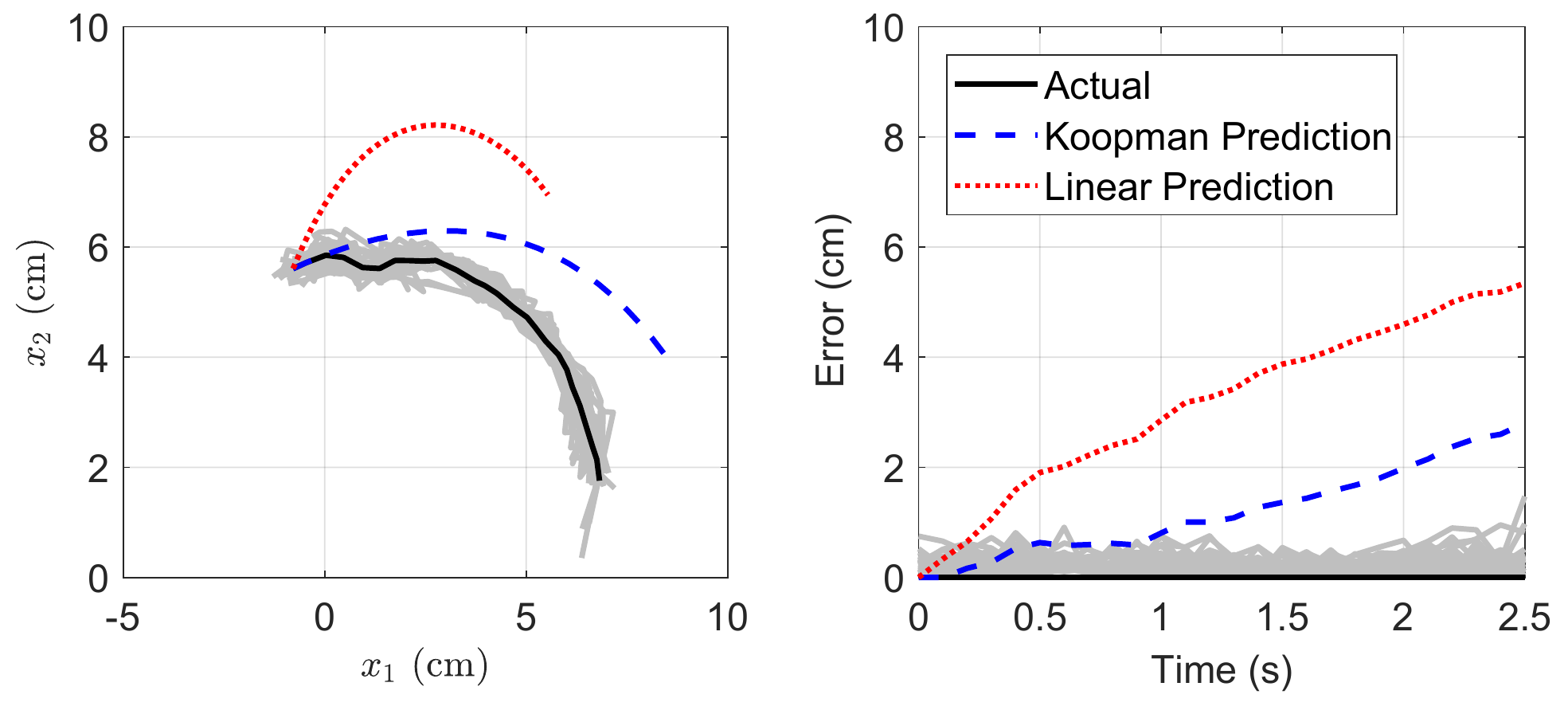}
    \caption{The (average) actual response and the model predictions for the robot over a 2.5 second horizon with the sinusoidal inputs described in \eqref{eq:unoise} with period $T = 10$ seconds applied. The left plot shows the actual trajectory of the laser dot along with model predictions. The error displayed on the right plot is defined as the Euclidean distance between the predicted laser dot position and the actual position at each point in time. The prediction error is smaller for the Koopman model over the entire horizon.}
    \label{fig:predict}
\end{figure}

The accuracy of the predictions generated by each of the two models were evaluated by comparing them to the actual behavior of the system under the sinusoidal inputs defined in \eqref{eq:unoise}.
The model responses were simulated over a time horizon of $2.5$ seconds given the same initial condition and input as the real system.
The results of this comparison are summarized by Fig. \ref{fig:predict} and Table \ref{tab:predict}. 
They illustrate that the Koopman model predictions are more accurate over the time horizon.

\begin{table}
    \rowcolors{2}{white}{gray!25}
    \setlength\tabcolsep{5pt} 
    \centering
    \caption{Average Prediction Error Over 2.5 second Horizon (cm)}
    \begin{tabular}{|c|c|c|c|c|c|c|c|c|}
        \hline
        \rowcolor{white} 
        & \multicolumn{7}{c |}{\textbf{Period of Sinusoidal Inputs (seconds)}} & \\
        \cline{2-8} \rowcolor{white}
        \multirow{-2}{*}{\textbf{Model}} & 6 & 7 & 8 & 9 & 10 & 11 & 12 &\multirow{-2}{*}{\textbf{Avg.}} \\
        \hline
        Koopman &  2.21  &  2.78 &  1.35  &  1.53  &  1.21 & 0.66 & 1.41 & 1.59 \\
        Linear S.S.  &  4.64  &  4.54  &  3.94 &  3.56  & 3.15 & 2.72 & 2.83 & 3.63 \\
        \hline
    \end{tabular}
    \label{tab:predict}
\end{table}

\subsection{Experiment 2: Model-Based Control Comparison}
\label{sec:mpcexp}

\begin{figure*}
    \centering
    \includegraphics[width=\linewidth]{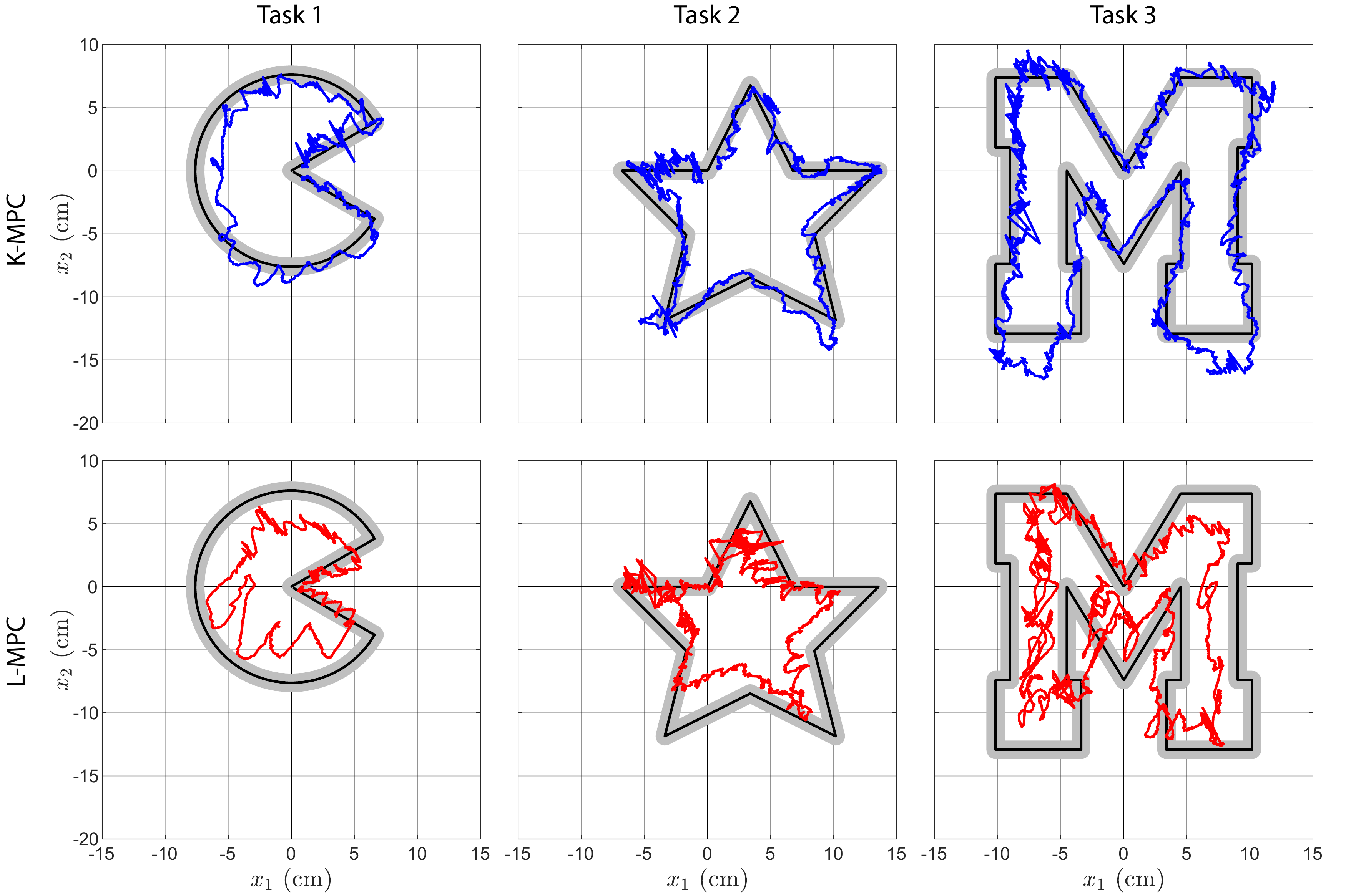}
    \caption{The results of the K-MPC controller (row 1, blue) and the L-MPC controller (row 2, red) in performing trajectory following tasks 1-3. The reference trajectory for each task is subimposed in black as well as a grey buffer with width equal to two standard deviations of the noise probability density shown in Fig. \ref{fig:noise}.}
    \label{fig:results}
\end{figure*}

The two identified models were each used to build a model predictive controller which solves an optimization problem in the form of \eqref{eq:mpc} at each time step using the Gurobi Optimization software \cite{gurobi}.
We refer to the two controllers by the abbreviations K-MPC for the one based on the Koopman model, and L-MPC for the one based on the linear state-space model. 
Both model predictive controllers run in closed-loop at $10$ Hz, feature an MPC horizon of 2.5 seconds ($N_h = 25$), and a cost function that penalizes deviations from a reference trajectory $r[k]$ over the horizon with both a running and terminal cost:
\begin{align}
    \text{Cost} &= 100 \left( y[N_h] - r[N_h] \right)^2 + \sum_{i=0}^{N_h - 1} 0.1 \left( y[i] - r[i] \right)^2
\end{align}
In the K-MPC case, $y[i] = C z[i]$, where $C$ is defined as in \eqref{eq:C}.
In the L-MPC case, $y[i] = C_L x_L[i]$ where $x_L$ is the four dimensional system state and $C_L$ is the projection matrix that isolates the states describing the current laser dot coordinates. 


The performance of the controllers was assessed with respect to a set of three trajectory following tasks.
Each task was to follow a reference trajectory as it traced out one of the following shapes over a certain amount of time:
\begin{enumerate}
    \item Pacman (90 seconds)
    \item Star (180 seconds)
    \item Block letter M (300 seconds)
\end{enumerate}
The error for each trial was quantified as the Euclidean distance from the reference trajectory at each time step over the length of the trial.

The performances of the K-MPC and L-MPC controllers at Tasks 1, 2, and 3 are shown visually in Fig. \ref{fig:results}, and the error is quantified in Table \ref{tab:RMSE}.
In both tasks the K-MPC controller achieved better performance, exhibiting an average tracking error of 1.26~cm compared to the L-MPC controller's average error of 2.45~cm.
This amounts to an average error roughly $25\%$ larger than than the maximum magnitude of observed noise (see Fig. \ref{fig:noise})

\begin{table}
    \rowcolors{2}{white}{gray!25}
    \setlength\tabcolsep{5pt} 
    \centering
    \caption{Average Error in Trajectory Following Tasks (cm)}
    \begin{tabular}{|c|c|c|c|c|c|}
        \hline
        \rowcolor{white} 
        & \multicolumn{3}{c |}{\textbf{Task}} & & \textbf{Std.} \\
        \cline{2-4} \rowcolor{white}
        \multirow{-2}{*}{\textbf{Controller}} & $1$ & $2$ & $3$ & \multirow{-2}{*}{\textbf{Avg.}} & \textbf{Dev.} \\
        \hline
        K-MPC &  1.25  &  1.19 &  1.34  &  1.26  &  0.07  \\
        L-MPC  &  2.21  &  2.34  &  2.73 &  2.45  &  0.31   \\
        \hline
    \end{tabular}
    \label{tab:RMSE}
\end{table}

\section{Conclusion}
\label{sec:conclusion}

In this work, a data-driven modeling and control method based on Koopman operator theory was successfully applied to a soft robot.
The Koopman-based MPC controller was shown to be capable of commanding a soft robot to follow a reference trajectory better than an MPC controller based on another linear data-driven model.
By making explicit control-oriented models of soft robots easier to construct, this method enables the rapid development of new control strategies and applications.

While these preliminary results are promising, further work is needed to make such methods feasible for higher dimensional robotic systems.
Toward that end, this work introduced a method for promoting sparsity in matrix representations of the Koopman model.
Additional work will explore strategies for further promoting sparsity, choosing the most effective basis of observables, and building models that can account for external loading and contact forces.





\section*{Acknowledgments}
This material is supported by the Toyota Research Institute, and is based upon work supported by the National Science Foundation Graduate Research Fellowship Program under Grant No. 1256260 DGE. Any opinions, findings, and conclusions or recommendations expressed in this material are those of the author(s) and do not necessarily reflect the views of the National Science Foundation.


\bibliographystyle{plainnat}
\bibliography{references}

\end{document}